# Modeling of New Energy Vehicles' Impact on Urban Ecology Focusing on Behavior

Run-Xuan Tang[1][0009-0007-8664-6505]

[1] Beijing Jiaotong University, Beijing, China
22241101@bjtu.edu.cn

**Abstract.** The surging demand for new energy vehicles is driven by the imperative to conserve energy, reduce emissions, and enhance the ecological ambiance. By conducting behavioral analysis and mining usage patterns of new energy vehicles, particular patterns can be identified. For instance, overloading the battery, operating with low battery power, and driving at excessive speeds can all detrimentally affect the battery's performance. To assess the impact of such driving behavior on the urban ecology, an environmental computational modeling method has been proposed to simulate the interaction between new energy vehicles and the environment. To extend the time series data of the vehicle's entire life cycle and the ecological environment within the model sequence data, LSTM model with Bayesian optimizer is utilized for simulation. The analysis revealed the detrimental effects of poor driving behavior on the environment.

## 1    Introduction

In promoting new energy vehicles and hybrids, while the 19% global market share in 2022 positively impacts environmental protection and diminishes fossil fuel consumption, concerns regarding the battery's environmental effects are increasingly prevalent and represent a significant environmental challenge.[1] The technological restrictions generated by batteries encompass vital environmental issues such as the strain exerted on the grid by excessive energy demand [2] and battery material recycling [3].

Mousaei [4] argues that the management of battery temperature must not be overlooked when evaluating energy management for hybrid electric vehicles (HEVs), as it significantly affects battery longevity. The Sitapure and Kulkarni study [5] highlights the importance of predicting the state of the battery, including charge and temperature, in creating an advanced battery management system (BMS).

Additionally, literature indicates that data-driven time series modeling was developed to predict battery lifetime, showing the influence of environmental factors and integrated parameters on battery performance. Cui et al. [6] conducted a study comparing new methods for EV charging and battery energy conditioning. They showed that enhancing the power handling system of aging battery packs can significantly improve the performance of a used battery storage system (2-BESS). Another study by Alexeenko and Charisopoulos [7] demonstrated that utilizing a shared battery resource can substantially reduce the deployed battery capacity, which has the potential to improve battery efficiency.

For the issue of driving behavior in new energy vehicles, Gebbran et al. [8] highlighted the significance of minimizing losses and optimizing path planning when analyzing dynamic electric vehicle path planning that takes battery degradation into account. Additionally, Herle, Channegowda, and Prabhu [9] enhanced the effectiveness of machine learning algorithms in predicting battery state by utilizing deep neural networks to address the issue of limited battery data.

In the study conducted by Schwenk and colleagues [10], they integrated battery aging to the two-way charging optimization process for electric vehicles (EVs). This underlines the significance of considering the effects of battery aging on operational expenses and the futuristic potential of intelligent charging applications for EVs. Furthermore, for the problem of routing electric vehicle batteries, Rodríguez-Esparza and his colleagues [3] proposed a hyper-heuristic algorithm that combines adaptive simulated annealing and reinforcement learning. This method could offer fresh insights into resolving power range restrictions and mitigating travel time and expenses.

A synthesis of the research presented above leads to the following two foci:

1. Quantitative assessment of the environmental impact of new energy vehicle driving behavior.
2. Exploration of the mechanism of the influence of new energy vehicle user behavior on the realization of expected environmental benefits.

For the first focus, it is evident that the driving habits of new energy vehicles have a direct effect on the environment, based on previous research. For the second focus, this study integrates state parameters of new energy vehicles and driver behavior data in simulating urban ecosystems computationally. An LSTM model by utilizing a Bayesian optimization data-driven approach [9] to examine environmental damages brought about by driver behavior patterns. This study also conducts significance tests to understand the long-term impacts in more detail. Furthermore, a significance test is carried out to obtain a more detailed comprehension of the long-term effects.

## 2     Related work: Environmental benefits of new energy vehicles

This paper focus on the behavior of new energy vehicle drivers. Battery optimization techniques and power network structures fall outside the scope of our study. As a key measure to reduce pollution emissions, electric vehicles are deemed a promising option [1]. It is crucial to note, however, that the environmental impact of electric vehicles is closely tied to the power supply. The type of energy utilized, such as coal, nuclear, or hydroelectric power, greatly affects the environmental impact of the power supply [3].

According to a life cycle assessment of electric vehicle batteries [14], traveling the same distance with electric vehicles equipped with different battery types results in significantly varying environmental impacts. Lithium-sulfur batteries are deemed the most appropriate option due to their cleaner properties throughout their life cycle. However, a significant number of electric vehicle (EV) users tend to disregard the practical lifespan of their EVs. This negligent behavior results in the abandonment of used batteries in the wild, giving rise to probable environmental hazards [16]. Inadequate disposal of these batteries can inflict a long-term catastrophic impact on the ecosystem, detrimentally affecting plant and animal species, and even posing a threat to human health [17].

Excluding battery recycling and disposal concerns, the scarcity of charging infrastructure is a significant hurdle for the electric vehicle industry. Chen et al. [11]found that the absence of charging stations increases consumer anxiety regarding mileage. Furthermore, studies suggest that unmanaged user behavior could hinder the potential environmental advantages of advanced electric vehicle technology [12].

This paper examines the psychological behavior of electric vehicle users. It discusses the impact of people's perceptions, use paradoxes, range anxiety, and gender attitudes on ownership [19]. The actual use of electric vehicles and consumer purchasing decisions are crucial for achieving their carbon emission benefits [18].

In addition, it is crucial to acknowledge the impact of diverse consumer behaviors and control group vehicle inventories to achieve the emission benefits of electric vehicles [13]. Consequently, assessing the effect of consumer behavior diversity on the realization of environmental advantages of EVs provides valuable insight for developing effective EV purchase incentives in the future [16].

The environmental impact of electric vehicle (EV) battery storage requires thorough investigation. Battery performance declines with time, particularly at high states of charge, which can worsen battery performance and impact its longevity and environmental efficiency [14]. However, little research has examined this matter from the viewpoint of EV owners, creating ambiguity surrounding the application of the term "high state of charge" in practice.

Amidst the increasing global consensus on environmental protection and emissions reduction, this research highlights the significance of driver behavioral improvements, instead of focusing solely on technical optimization, to fully comprehend and actualize the potential of electric vehicles (EVs) [3].

## 3  Insight of Vehicle Status and Driver Behavior

Current research recognizes common predictions, such as the gradual phasing out of fossil-fueled power plants, lowering emissions during electric vehicle (EV) production, and changes to the automotive energy consumption trend. These forecasts are useful in predicting the future, but they carry the risk of deviating from reality [21]. Our investigation encompasses both renewable and fossil fuel power sources by including electric vehicles and hybrid electric vehicles (HEVs).

To redefine vehicle behavior analysis, a framework that incorporates temporal, spatial, and energy dimensions is developed. The framework aims to provide an objective and comprehensive analysis of vehicle behavior. This data-centric technique aggregates multidimensional data streams which can be mined for insightful characteristics of electric vehicle functionality.

Analyzing multiple types of vehicles, specifically personal electric vehicles, across a variety of temporal and urban classifications reveals intricate driver behavior patterns, charging habits, and vehicle usage [20].

Using this analysis, a detailed investigation into electric vehicle travel data is conducted to comprehend users' driving characteristics thoroughly. This is augmented by ample data from charging stations, facilitating the recognition of electric vehicle (EV) charging dynamics, encompassing time and power metrics that serve as fundamental components for comprehending user charging behavior.

With these fundamental elements in place, the relationship between driving inclinations and charging patterns can be researched. This study on traffic behavior, trip mapping, and charging tendencies offers valuable insights into EV usage from the standpoint of owners.

The process entailed several stages of analysis on vehicle data, initiated with transforming time fields into a timestamp format, which facilitated subsequent aggregation and analysis tasks. Subsequently, a state change analysis was conducted, during which shifts in vehicle states, such as startups and shutdowns, were examined over time in order to comprehend usage patterns. Furthermore, the behavior of the charging mode was also analyzed over time in order to discern the prevalence and effectiveness of different charging strategies, including during driving and parked scenarios.

The operation mode analysis encompassed two principal elements: the distribution of operational modes, such as pure electric versus hybrid usage, and an assessment of driving behaviors corresponding to these modes; and mode switching analysis, which aimed to identify prevalent switching tendencies among vehicles and the conditions prompting such switches.

Outlier detection procedures were implemented in three dimensions: temporal anomalies that flagged irregularities in timestamps, state anomalies detecting unconventional vehicle or charging status changes, such as immediate shutdown post-startup, and mode anomalies which highlighted inconsistencies in operation mode shifts, for instance, transitioning to fuel mode during a charging session. Moreover, feature-based analyses of driver behavior were conducted, with driving and charging modes serving as bases. These included assessments of whether drivers who frequently opt for pure electric mode exhibit more eco-friendly practices and of the circumstances under which drivers choose to utilize driving as a means to charge the vehicle.

Correlative insights between the state of charge (SOC) of the electric vehicle over time and several pivotal indicators were visually presented through correlation analyses, encapsulated in two diagrams. These diagrams further enrich the understanding of the interplay between SOC management and overall vehicle performance.

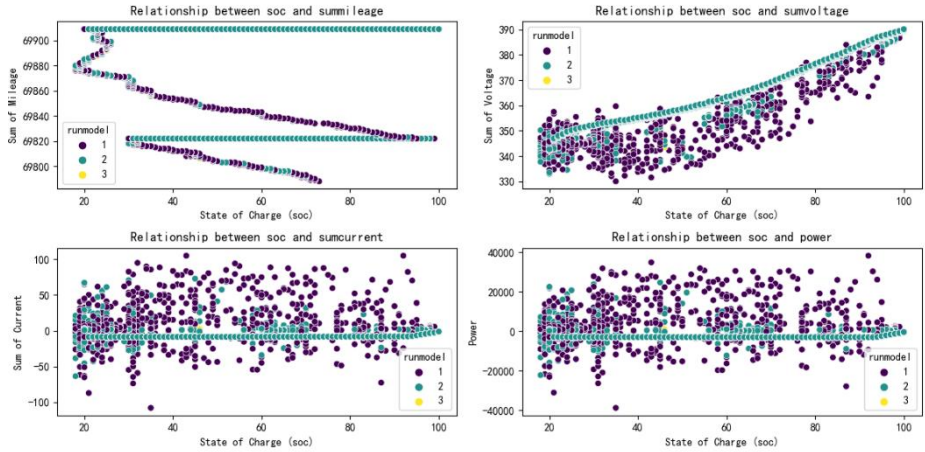

**Fig. 1.** Relationship between soc and sum-mileage, sum-voltage, sum-current, power

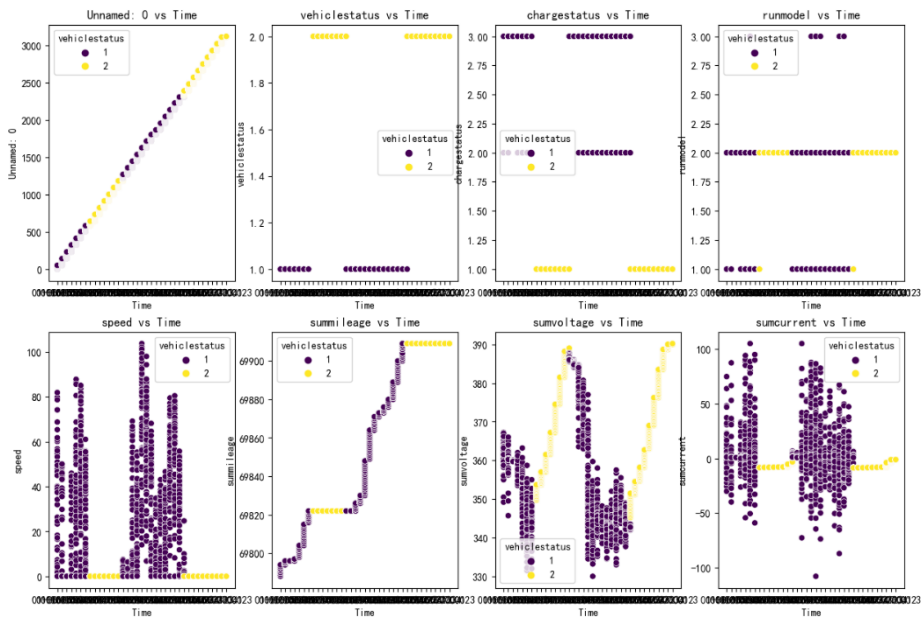

**Fig. 2.** Data analysis for hybrid vehicles

There is a positive correlation between the state of charge (SOC) and the cumulative mileage (sum of mileage) of the vehicle, suggesting that higher battery SOC could lead to greater mileage.

Additionally, there is a positive correlation between SOC and the sum of voltage, indicating a relationship between battery SOC and voltage.

A negative correlation exists between SOC and battery current. This implies that when the battery's state of charge is high, the battery current is low. And, a negative correlation exists between SOC and vehicle power. This means that when the battery's state of charge is high, the vehicle power decreases.

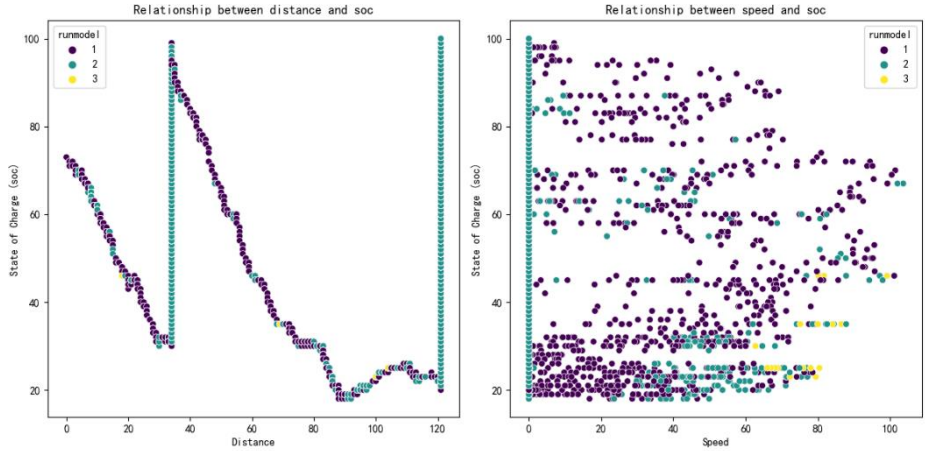

**Fig. 3.** The relationship between soc and distance, soc and speed

A negative correlation exists between distance and SOC, suggesting that the battery's charge decreases as the vehicle travels a greater distance. Besides, a negative correlation exists between speed and SOC, indicating that the battery's charge decreases as the vehicle travels faster.

## 4       Urban Ecosystem Modeling

An all-encompassing assessment model has been established to evaluate the ecological effects of land degradation, pollutant concentrations, climate change, forest rehabilitation, waste management, biodiversity, levels of industrialization, traffic expansion, energy usage, new energy transportation, and the chemical toxicity of lithium-sulfur batteries.

The model aims to examine the use of lithium-sulfur batteries, their chemical toxicity, and the environmental impact of carbon emissions resulting from the utilization of electric energy in new energy cars. Collect and process data using appropriate pre-processing and feature engineering techniques. To incorporate new energy vehicles into this ecosystem model, the electrical consumption by these vehicles was converted to carbon emissions, following international standards. Harmful substances produced in battery production were also quantified, using international standards.

An LSTM was trained using a Bayesian parameter optimizer, and this LSTM was used to extend all data in the time dimension. All weight factors in this model were acquired through the information Entropy Weighting Method (EWM). The model illustrates the relationship between new energy vehicles and the urban environment through the use of simulation modeling. These indicators and interactions are expressed through the following mathematical equations.

## 4.1 Data Extending

The following figures show how the LSTM works with our data.

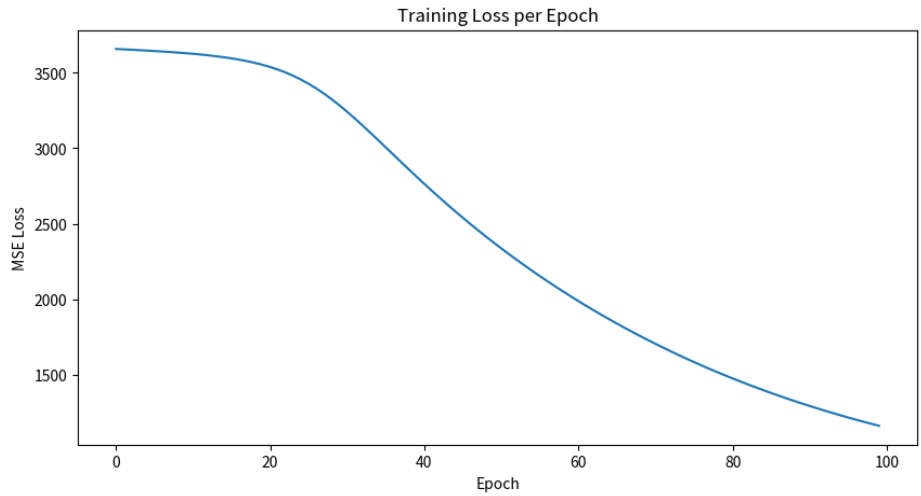

**Fig. 4.** Decline curve of training loss

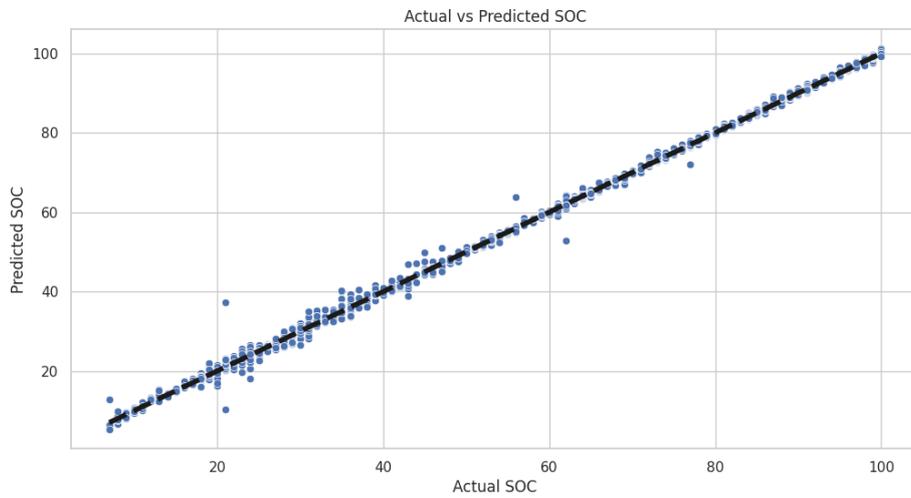

**Fig. 5.** Test goodness of fit

**Table 1.** Evaluation of LSTM with Bayes-Optimizer

|  | Metric | Value |
|---|---|---|
| R2 | Train | 0.999274 |
|  | Test | 0.998679 |

|     | Train | 0.497713 |
| --- | --- | --- |
| MSE | Test | 0.938392 |

Table 1 indicates the model exhibits comparable goodness of fit during training and testing. However, the mean squared error (MSE) during training is approximately half that observed during testing. This may suggest that the model is exhibiting signs of overfitting.

### 4.2  Simulating

The evaluation indicators are summarized [23,24]. All the $\epsilon$ symbols below denote random error terms.

**Land Degradation (LD).**

$$LD = l_1 \cdot Pop + l_2 \cdot T_r + l_3 \cdot \alpha + l_4 \cdot A + \epsilon_{LD} \quad (1)$$

($Pop$): Population density.
($T_r$): Rate of change of regional temperature.
($\alpha$): Parameter of land use pattern, reflecting the type and intensity of land use.
(A): The size of the area that may influence land degradation.

**Pollutant Concentration (PC).**

$$PC = g_1 \cdot C_{industrial} + g_2 \cdot C_{traffic} + g_3 \cdot C_{waste} + g_4 \cdot RAD + \epsilon_{PC} \quad (2)$$

($C_{industrial}$): Industrial pollutant concentration.
($C_{traffic}$): Concentration of pollutants caused by traffic.
($C_{waste}$): Concentration of pollutants caused by waste.
($RAD$): Residential Area Disturbance Degree (RAD): The impact of residential activities on the environment.

**Climate Change Impact (CCI).**

$$CCI = c_1 \cdot \Delta Temperature + c_2 \cdot \Delta Precipitation + \epsilon_{CCI} \quad (3)$$

($\Delta Temperature$): Temperature difference.
($\Delta Precipitation$): Precipitation difference.

**Forest Restoration (FR).**

$$FR = d_1 \cdot VFC + d_2 \cdot DI + d_3 \cdot R + \epsilon_{FR} \quad (4)$$

($VFC$): Vegetation cover.

($DI$): Forest Destruction Index.
($R$): Forest recovery rate

**Waste Treatment Rate (WTR).**

$$WTR = e_1 \cdot C_{\text{waste}} + e_2 \cdot P + \epsilon_{WTR} \qquad (5)$$

($C_{waste}$): Concentration of waste generated.
(P): Waste treatment policy and treatment efficiency.

**Lithium Sulfur Battery Chemical Toxicity (LSBCT).**

$$LSBCT = g_1 \cdot Charging\ Behavior + g_2 \cdot Temperature + g_3 \cdot Driving\ Speed + \epsilon_{LSCBT} \quad (6)$$

In the Lithium Sulfur Battery Chemical Toxicity (LSBCT) equation, the function (p) establishes a relationship between charging behavior, temperature, and driving speed. This function serves as a model for evaluating the potential environmental impacts resulting from improper use of lithium-sulfur batteries.

Ultimately, these impacts are assigned weight values and added together to calculate the NEV Efficiency.

**NEV Efficiency.**

$$E = w_1 \cdot LD + w_2 \cdot PL + w_3 \cdot CCI + w_4 \cdot FR + w_5 \cdot WTR + w_6 \cdot NEVI + w_7 \cdot LSBCT + \epsilon_E \quad (7)$$

The above model highlights that overcharging and discharging, as well as the habit of frequent recharging only when the battery is very low, significantly increase the chemical toxicity emissions from lithium-sulfur batteries.

Furthermore, operating a lithium-sulfur battery at high speeds can cause overheating and a higher risk of harmful chemical emissions.

**Parameters Setting.**

$$NEV\ Population = 31.6\% \qquad (8)$$

The heightened chemical toxicity of lithium-sulfur batteries not only harms battery performance but also negatively impacts the environment, increasing the urgency for implementing technology and modifying behaviors to mitigate traffic-related pollution.

**Table 2.** Results of Environmental Impact

| Indicator | Before Simulation | After Simulation |
|---|---|---|
| Land Degradation (LD) | 3.22 | 3.53 |
| Pollutant Concentration (PC) - SO2 | 0.04 mg/m³ | 0.07 mg/m³ |
| Climate Change Impact (CCI) | 0.21° C | 0.32° C |
| Forest Restoration (FR) | 0.32% | 0.26% |
| Waste Treatment Rate (WTR) | 20.3% | 15.5% |
| Lithium Sulfur Battery Chemical Toxicity (LSBCT) | 4 | 4 |

Table 3. Results of NEV Efficiency

| Indicator | Values | Standardization |
|---|---|---|
| NEV Growth(ΔNEV Population) | 0.001321 | 1 |
| ΔNEV Efficiency | -0.004252 | -3.21877 |

## 5  Discussion

During our analysis and mining of locomotive state and driver behavior data, it was observed that electric vehicles exhibited phenomena such as high-speed driving and battery overloading. Based on the correlation between the state of charge (SOC) and related indexes, an increase in driving speed leads to a reduction in battery charge, an increase in battery power, and subsequently results in overheating of the batteries, causing melting and production of toxic substances. These actions shorten battery life and adversely affect the environment. Charging the battery only until it reaches its low limit or constantly fully charging it can harm the battery's performance and reduce its lifespan.

NEV Growth, which is 0.001321, indicates that an increase in the per-unit popularity of new energy vehicles will have a slightly positive impact on the environment, specifically increasing landscape diversity and reducing disturbance in residential areas.

However, enhancements in NEV efficiency do not necessarily result in expected environmental benefits.

According to NEV Efficiency which is -0.004246, technological advancements may not offset the negative environmental effects of inappropriate vehicle use habits. Thus, improved efficiency does not inevitably lead to environmental benefits.

The driving habits of electric and hybrid vehicle operators suggest that they tend to overcharge and overuse electricity, resulting in damage to the battery and the surrounding ecosystem. By applying long short-term memory (LSTM) to the timeseries data of automobiles and extending it to the end of their lifecycle, this information is integrated into our ecological evaluation model calculations. The outcome of this process revealed that the implementation of new energy vehicles does not significantly enhance the environment, thus indicating that these vehicles do not fulfill the anticipated goals of effective energy conservation and emissions reduction. This study suggests that new energy vehicles do not meet the expectations of effectively saving energy and reducing emissions.

The research primarily focuses on vehicle owners' behavioral patterns to understand this phenomenon. Furthermore, current research has not determined the optimal use of lithium-sulfur batteries, which are commonly used in new energy vehicles. This is a crucial area for further investigation. In addition, future policies must be implemented to improve hardware and software related to monitoring driver speed. Finally, a significant advancement is made in computationally analyzing the interaction between electric vehicles and the environment with urban ecosystem modeling. These findings hold substantial importance.

## 6      Conclusion

In this study, an analysis is conducted on the alteration of parameters concerning two emerging energy vehicles, namely electric and hybrid vehicles, in conjunction with driver behavior. The research shows that certain driver behavior such as overcharging, discharging, and excessive speed have the potential to be detrimental to the battery, thereby posing a risk to the environment. Based on our findings, a simple yet effective computational simulation model is proposed for urban ecosystems. It covers data generation, evaluation, and analysis. The results of our simulations confirm that poor driving habits have adverse effects on urban ecology. The work encourages relevant authorities to improve driving behaviors. Additionally, the urban ecosystem simulation model offers an excellent testing environment for simulation experiments.